\documentclass[11pt]{article}

\usepackage[preprint]{acl}

\usepackage{times}
\usepackage{latexsym}

\usepackage[T1]{fontenc}
\usepackage[utf8]{inputenc}
\usepackage{microtype}
\usepackage{graphicx}
\usepackage{amsmath}
\usepackage{amssymb}
\usepackage{booktabs}
\usepackage{tabularx}
\usepackage{float}

\title{RexBERT: Domain Adapted Encoders for Product Understanding and Relevance Ranking in E-commerce}

\author{Rahul Bajaj\thanks{Corresponding author.}\thanks{Primary contributor.} \\
  \texttt{thebajajra@gmail.com} \\
  \And
  Anuj Kumar Garg\footnotemark[2] \\
  \texttt{anujgarg2004@gmail.com} \\
  \AND
  Jaya Nupur \\
  \texttt{thejayanupur@gmail.com} \\
  \And
  Qixing Wang \\
  \texttt{qxbwang@gmail.com} \\
}
\date{}
\begin{document}

\maketitle

\begin{abstract}
E-commerce retrieval and recommendation require encoders that preserve
product specific distinctions, including substitutes versus complements,
attribute equivalence, brand/category constraints, and query product intent.
General purpose encoders trained on broad web text often underrepresent these
fine grained compatibility and specification signals.  We introduce RexBERT, a
family of encoder models with 17M to 400M parameters for e-commerce product
understanding.  Starting from the public Ettin long context encoder, RexBERT
is domain annealed on Ecom-niverse, a 350B-token e-commerce domain language corpus,
while retaining an 8{,}192-token context window.  We further introduce Guided
MLM, a masking strategy that preferentially selects salient product spans
without changing the masked language modeling objective.  We evaluate RexBERT
on e-commerce domain language modeling, ESCI graded query product similarity,
WANDS relevance ranking, and GLUE retention checks.  Domain annealing
substantially improves held out e-commerce modeling: at Base scale, perplexity
drops from 6.04 to 3.99 after Phase~3 training with Guided MLM.  On Amazon
ESCI, RexBERT-base already matches E5-large with fewer than half the
parameters, while RexBERT-large reaches 0.457 Spearman
correlation, outperforming E5-large and BGE-large under the same graded
similarity protocol.  On WANDS, RexBERT-large achieves 0.924 NDCG@5, 0.933
NDCG@10, and 0.977 MAP, while maintaining competitive GLUE retention.  These
results show that targeted domain annealing and guided span masking can deliver
stronger e-commerce relevance than general purpose encoders at compact encoder
scales.
\end{abstract}

\section{Introduction}

Encoder only language models remain central to production retrieval and
reranking because they offer a favorable tradeoff between quality, latency and
cost.  BERT~\citep{devlin2019bert} and its successors support high throughput
matching systems, while recent work such as
ModernBERT~\citep{warner2025smarter} shows that encoders can still improve
through better recipes, longer contexts and more efficient attention.  The open
question we study is whether such encoders can be adapted to the
product centric semantic relations that determine e-commerce relevance.

E-commerce semantics differ from generic sentence similarity in systematic
ways.  Queries are often short and underspecified, product text is
semi structured, and small attributes such as brand, size, color, material,
quantity or compatibility can determine whether two items are interchangeable.
For example, ``iPhone 14 case'' and ``iPhone 14 phone'' share many tokens but
refer to complementary rather than substitutable products, while ``women's
waterproof trail running shoes size 8'' depends on gender, use case, material
and size constraints.  Generic encoders trained on broad web corpora often blur
these distinctions, especially for long tail entities and attribute value
constructions.

Our approach combines Ecom-niverse, a domain annealing strategy, and
Guided MLM (\S\ref{sec:guided_mlm}).  Given a strong long context checkpoint,
is a focused annealing phase enough to recover product specific semantics?  We
release RexBERT as pretrained encoders and evaluate them on downstream
benchmarks rather than pretraining loss alone.

\paragraph{Contributions} Our work makes the following contributions:
\begin{enumerate}
  \item We introduce \textbf{Ecom-niverse}, a 350B-token product language corpus
  curated from FineFineWeb~\citep{map2024finefineweb} through domain selection, LLM validation and
  classifier based filtering; and \textbf{RexBERT}, a family of compact
  e-commerce encoders domain adapted on this corpus, and release
  checkpoints and corpus data at the Hugging Face collection
  \url{https://huggingface.co/collections/thebajajra/rexbert}.
  \item We study Phase~3 domain annealing from a strong long context checkpoint
  as a practical e-commerce adaptation strategy (\S\ref{sec:training}).
  \item We propose \textbf{Guided MLM} and validate it with controlled
  perplexity ablations (\S\ref{sec:ablation_gmlm}).
  \item We evaluate RexBERT checkpoints against general domain and task matched
  baselines on product language, relevance, and general NLU benchmarks
  (\S\ref{sec:experiments}).
\end{enumerate}

\begin{figure*}[t]
  \centering
  \includegraphics[width=0.75\linewidth]{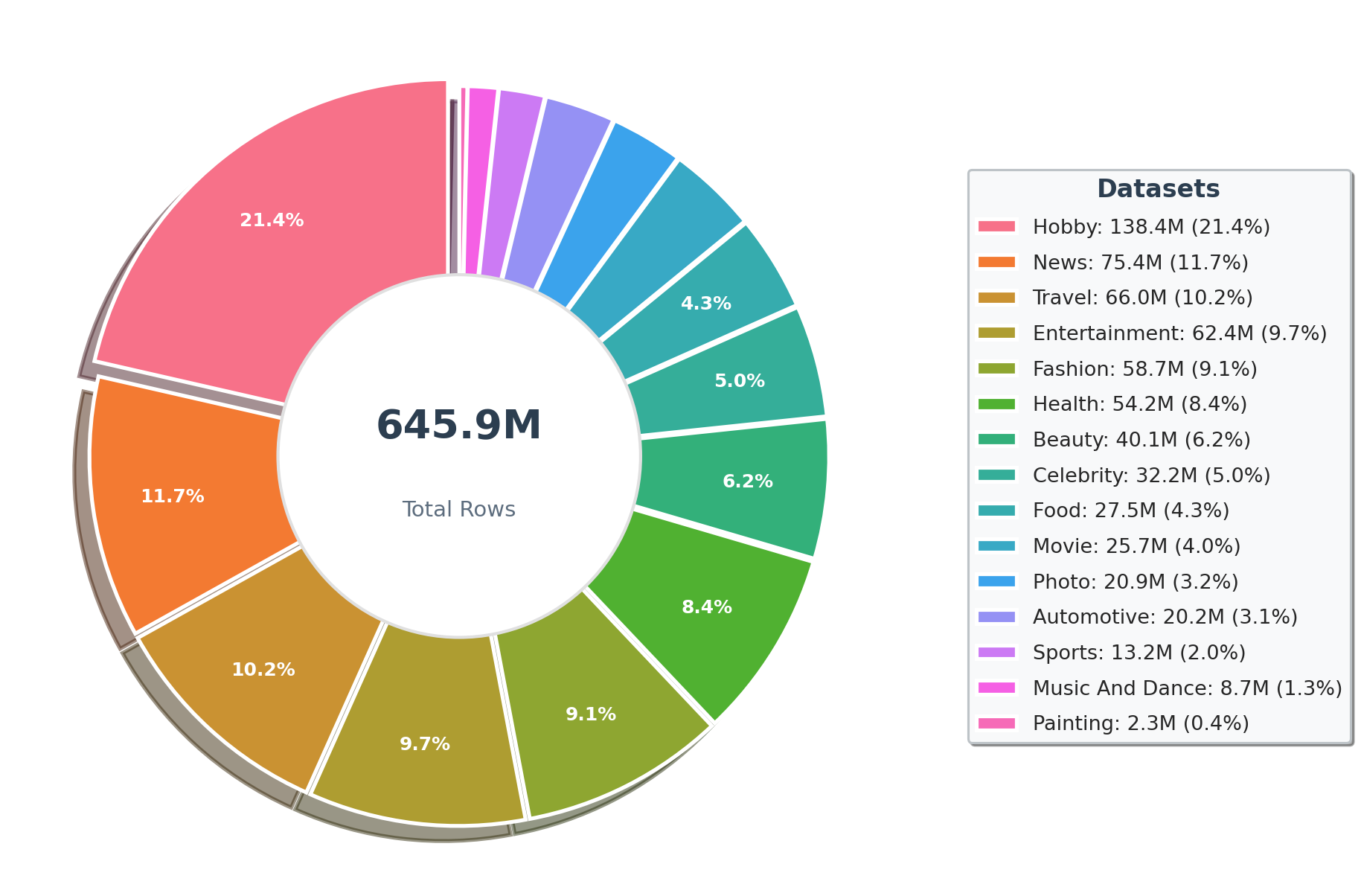}
  \caption{Domain distribution of Ecom-niverse.  Sizes represent the amount
  of filtered data contributed by each FineFineWeb domain and additional
  corpora.  Hobby and News supply the largest portions.}
  \label{fig:domain_distribution}
\end{figure*}

\section{Related Work}
\label{sec:related}

\paragraph{Modern encoder pretraining.}  BERT~\citep{devlin2019bert} and
RoBERTa~\citep{liu2019roberta} established encoder only pretraining as a
strong foundation for classification and retrieval.  More recent work revisits
the encoder recipe itself: DeBERTa~\citep{he2020deberta} improves attention
parameterization, MosaicBERT~\citep{portes2023mosaicbert} and
Cramming~\citep{geiping2023cramming} study training efficiency, and
ModernBERT~\citep{warner2025smarter} combines RoPE, GeGLU, alternating
local/global attention, unpadding and flash attention for long context
encoders.  RexBERT adopts this modern encoder direction but asks whether the
training signal can be specialized for product centric semantic relations.

\paragraph{Domain adaptive encoders.}  Domain specific pretraining has improved
encoders in biomedical, clinical, scientific, financial and legal NLP, including
BioBERT~\citep{lee2020biobert}, ClinicalBERT~\citep{alsentzer2019publicly},
BioClinical ModernBERT~\citep{sounack2025bioclinical},
SciBERT~\citep{beltagy2019scibert}, FinBERT~\citep{yang2020finbert} and
LEGAL-BERT~\citep{chalkidis2020legal}.  These studies show that targeted
corpora can improve downstream behavior, but evaluation typically targets
domain specific classification or tagging rather than product retrieval.
E-commerce search additionally requires graded query product relevance and
attribute sensitive matching.

\paragraph{E-commerce and product representation.}  E-BERT augments BERT with
phrase level and product level pretraining tasks for e-commerce
\citep{zhang2020bert}, while CatBERT trains incrementally on catalog text and
argues that product data differs sharply from sentence like corpora
\citep{mallavarapu2022catbert}.  RexBERT follows this domain specialization
motivation but emphasizes open data, compact encoder scales,
and Guided MLM as an attribute aware masking mechanism.

\paragraph{Retrieval, reranking and product benchmarks.}  Query product matching
is fundamentally a retrieval and ranking problem rather than generic sentence pair
classification.  Because of this framing, standard evaluations compare dense
embedding models such as
E5~\citep{wang2022text}, BGE~\citep{xiao2024c}, and
GTE~\citep{li2023towards}, cross encoder rerankers~\citep{nogueira2019passage}, and
general encoders including ModernBERT~\citep{warner2025smarter}.  For
e-commerce, Amazon ESCI~\citep{reddy2022shopping} provides graded
query product similarity labels and WANDS~\citep{chen2022wands} evaluates
query level ranking over human relevance judgments.

\section{Data and Corpus Curation}
\label{sec:data}

Fine grained domain representation is critical for accurate e-commerce
encoding.  We therefore curate \textbf{Ecom-niverse}, a large corpus of
commerce related text.  Our source is \emph{FineFineWeb}~\citep{map2024finefineweb}, a 4.4~trillion
token CommonCrawl derived corpus organised into fifty
categories.  Each entry consists of a text snippet and domain label.  To
isolate retail content we manually inspect domains and select those with
substantial overlap with commerce.  We group them into sources requiring
domain specific filtering and sources with near-complete commerce overlap
(Appendix~\ref{app:data_details}).  Notably,
the Hobby domain contributes 21.4\%, News contribute 11.7\%,
and Fashion and Beauty combined contribute 15.3\%.  Figure~\ref{fig:domain_distribution}
visualises the resulting distribution.

After selecting domains, we apply language detection to retain English text and
remove boilerplate, advertisements and low quality pages.  We deduplicate
near duplicate documents using MinHash and filter profanity and adult
content.  The resulting corpus comprises more than 350~billion tokens of
e-commerce text, providing a large scale domain specific resource for
pretraining.

\begin{figure*}[t]
  \centering
  \includegraphics[width=0.85\linewidth]{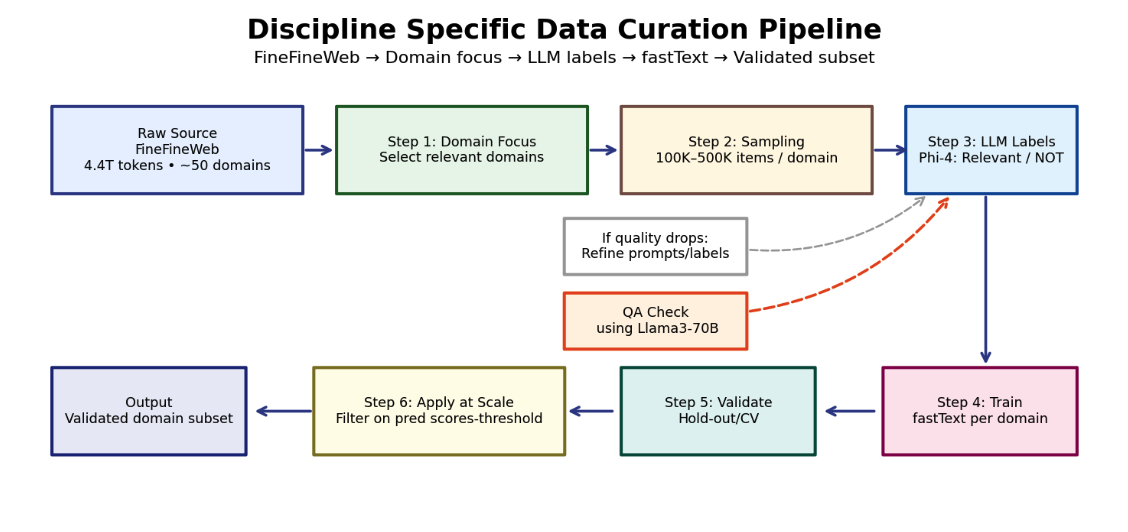}
  \caption{Ecom-niverse curation pipeline: domain selection, sampling, LLM labeling, QA auditing, fastText distillation, and thresholded filtering at scale.}
  \label{fig:data_pipeline}
\end{figure*}

\paragraph{Source corpus and cleaning pipeline.} After selecting
commerce overlapping FineFineWeb domains (Appendix~\ref{app:data_details}), we
sample examples from each for LLM assisted binary relevance labeling.  A stronger
auditor model reviews a subset of labels to identify prompt and guideline
failures.  The resulting labels are distilled into per domain fastText
classifiers, which score the full domain partitions at scale; domain specific
thresholds determine the retained pretraining subset.
Before Phase~3 training, we decontaminate the retained pretraining corpus
against the public evaluation benchmarks used in this paper, including Amazon
ESCI and WANDS.

\section{Pretraining}
\label{sec:training}

RexBERT continues from the public Ettin~\citep{weller2025seq} encoder
checkpoint after its
long context phase.  Phases~1 and 2 are inherited initialization
(Table~\ref{tab:training_provenance}); this paper's contribution is Phase~3
domain specialization on Ecom-niverse with annealed sampling and Guided MLM.
Our central question is how much e-commerce behavior Phase~3 alone can induce
without full domain pretraining from scratch.
Let $X=(x_1,\dots,x_T)$ denote a token sequence and $\theta$ the model
parameters.  The masked language modeling (MLM) objective is
\begin{equation}
\mathcal{L}_{\mathrm{MLM}}(\theta)
= -\,\mathbb{E}_{X\sim\mathcal{D},\, m\sim\mathcal{M}}
\big[\log p_\theta(X_{m}\mid X_{\setminus m})\big].
\end{equation}
where $m$ indexes masked positions drawn from a masking distribution $\mathcal{M}$ and $X_{\setminus m}$ denotes the unmasked context. Masking is performed dynamically with span aware sampling to better capture multi-token lexical units and to reduce information leakage from partially observed subword fragments.

\begin{table*}[t]
  \centering
  \small
  \caption{Training provenance and curriculum.  Phases 1 and 2 are inherited from
  the public Ettin long context encoder checkpoint; this paper performs the
  Phase~3 e-commerce domain specialization training.}
  \label{tab:training_provenance}
  \setlength{\tabcolsep}{4pt}
  \begin{tabular}{@{}llllll@{}}
    \toprule
    Phase & Starting checkpoint & Data & Tokens & Seq. length & Trained here? \\
    \midrule
    1: General pretraining & Ettin recipe/checkpoint lineage & Mixed open data & 1.7T & 512 & No \\
    2: Context extension & Ettin Phase~1 checkpoint & Mixed long context data & 250B & 8{,}192 & No \\
    3: Domain annealing & Ettin Phase~2 8{,}192-token checkpoint & Ecom-niverse & 350B & 8{,}192 & Yes \\
    \bottomrule
  \end{tabular}
\end{table*}

\subsection{Guided MLM}
\label{sec:guided_mlm}

Guided MLM is a targeted masking variant that prioritizes information rich
product spans.  Guidance comes from lightweight named entity recognizers: for
each FineFineWeb source domain we curate training labels with
GLiNER~\citep{zaratiana2024gliner} and fine-tune a BERT-mini tagger on
domain specific attribute and product entity annotations.  At Phase~3 training
time, sequences are tagged with the NER model matched to their source domain;
recognized spans highlight characteristic attributes and product features that
carry retrieval critical signal.  Let $p(s)$ denote the base span sampling
probability for candidate span $s$.  For tagged spans we set
$\tilde{p}(s)=\alpha\, p(s)$ and leave $\tilde{p}(s)=p(s)$ otherwise, then
renormalize within each sequence to obtain the guided masking distribution.
The MLM objective is unchanged.  During training, $5\%$ of sequences per batch
are masked under this guided distribution and the remainder use standard random
span masking.  This preserves the generalization benefits of unbiased MLM while
allocating additional learning signal to rare, high value tokens and structured
product descriptors.

\begin{table}[t]
  \centering
  \small
  \caption{Examples of Guided MLM span selection.}
  \label{tab:guided_mlm_examples}
  \begin{tabularx}{\linewidth}{@{}XX@{}}
    \toprule
    Text & Guided token spans \\
    \midrule
    Removable, washable cover Easy-carry handle Soft, warm and luxurious! Filled with reclaimed poly fiber to keep waste out of landfills. Denim adds a touch of softness to the rugged look of this comfortable pet bed. \#39103 blue denim wash & 
    Removable; washable; Easy-carry; reclaimed poly fiber; Denim; blue; \\
    \bottomrule
  \end{tabularx}
\end{table} 

\paragraph{Tokenizer and optimizer.}  RexBERT preserves the tokenizer and
long context architecture of the Ettin Phase~2 checkpoint.  The tokenizer is a
modern BPE vocabulary of 50{,}368 tokens, improving token efficiency over the
original BERT WordPiece; the inherited 8{,}192-token context window supports
long product text.  During our Phase~3 domain annealing, optimization uses
StableAdamW with Adafactor-style update clipping and a trapezoidal
Warmup, Stable, and Decay schedule with a final inverse square root
($1/\sqrt{t}$) decay; hardware, batch size, steps and seeds are omitted from
this submission.

Phases~1 and 2 inherit Ettin's 1.7T-token general pretraining and 250B-token
context extension to 8{,}192 tokens; RexBERT starts from the released Phase~2
checkpoint.

\begin{figure*}[t]
  \centering
  \includegraphics[width=0.75\linewidth]{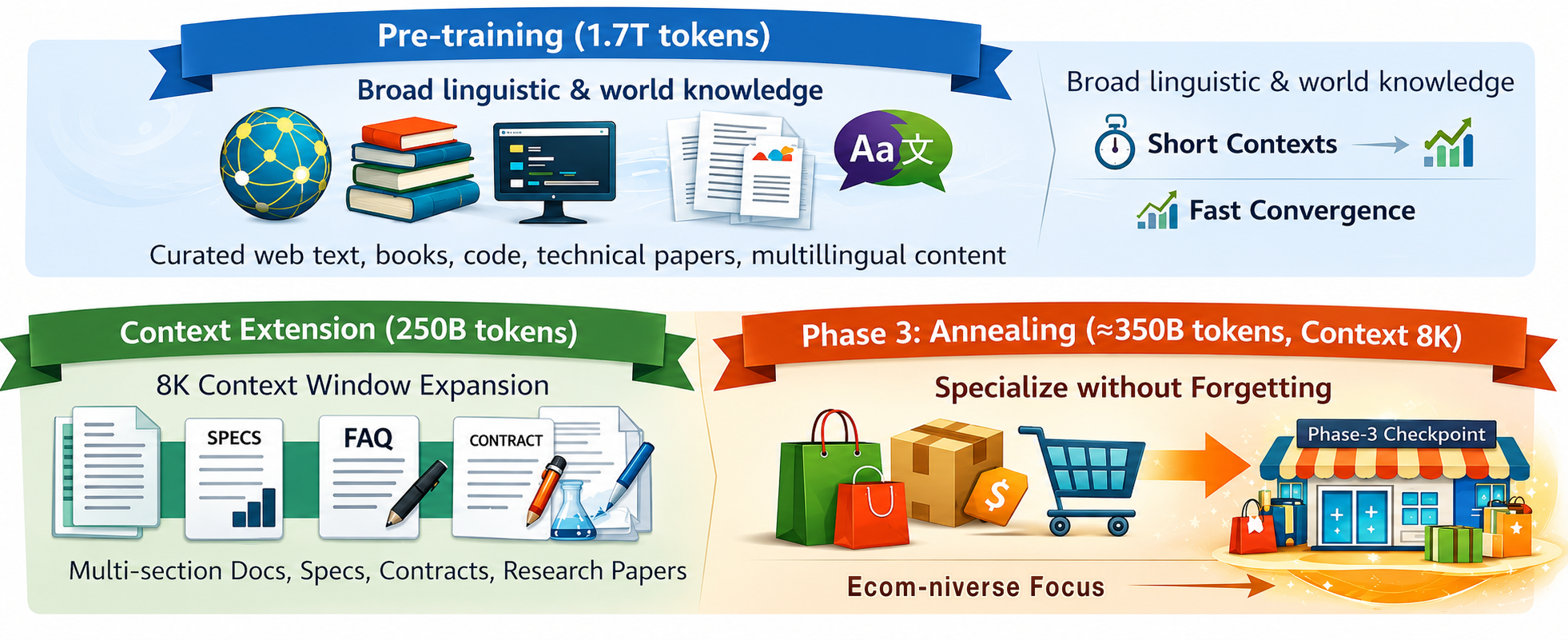}
  \caption{Training curriculum for RexBERT.  Phases 1 and 2 are inherited from
  the public Ettin long context encoder checkpoint; this work performs the
  Phase~3 annealing on Ecom-niverse for 350~billion tokens.}
  \label{fig:timeline}
\end{figure*}

\paragraph{E-commerce annealing.}  We specialize the Phase~2
checkpoint on Ecom-niverse for approximately 350~billion tokens.  We reduce the
masking ratio to 10\% to 15\%, anneal sampling weights toward e-commerce data, and
interleave Guided MLM with standard span masking.  The learning rate follows the
same Warmup, Stable, and Decay schedule with a final $1/\sqrt{t}$ decay while
preserving the inherited 8{,}192-token context (Figure~\ref{fig:timeline}).

\section{Model Architecture}
\label{sec:model}

RexBERT retains the BERT style encoder architecture and modern long context
enhancements inherited from the Ettin Phase~2 checkpoint.  Key differences from
the original BERT architecture include:
\begin{itemize}
  \item \textbf{Biasless layers and pre normalisation.}  Following
  ModernBERT, we remove bias terms from linear layers and layer norms and
  use pre layer normalisation to improve training stability.
  \item \textbf{Rotary positional embeddings.}  We replace absolute
  positional embeddings with RoPE, enabling extrapolation to long contexts
  and efficient implementation.
  \item \textbf{GeGLU activations.}  GeGLU activations provide better
  optimisation compared to GELU.
  \item \textbf{Alternating global/local attention.}  Attention layers
  alternate between full attention and local sliding windows, reducing
  quadratic complexity while preserving global context.
  \item \textbf{Unpadding and flash attention.}  We adopt unpadding to
  remove padding tokens and use Flash Attention for both global and local
  attention, improving throughput.
  \item \textbf{Hardware aware depth.}  The Micro, Mini, Base and Large
  variants follow a deep and narrow design to maximise GPU utilisation
  within parameter budgets.
\end{itemize}
Table~\ref{tab:sizes} in Appendix~\ref{app:model_configs} summarises the
configuration of RexBERT models.

\section{Experiments}
\label{sec:experiments}

We compare RexBERT checkpoints against general domain and task matched
baselines on downstream tasks (\S\ref{sec:semantic_similarity} and \S\ref{sec:ablation_gmlm}),
testing whether domain annealing transfers beyond pretraining perplexity.
We provide RexBERT checkpoints and Ecom-niverse artifacts for reproducibility at the Hugging Face collection
\url{https://huggingface.co/collections/thebajajra/rexbert}.

\paragraph{Datasets.}  We evaluate on Amazon ESCI~\citep{reddy2022shopping}
(English \texttt{small} split: 419{,}653 train / 181{,}701 test), WANDS~\citep{chen2022wands}
(140{,}068 / 46{,}690 / 46{,}690 train/dev/test), and GLUE~\citep{wang2018glue}
as a general NLU retention check.

\paragraph{Evaluation protocols.}  All models start from publicly available or
RexBERT pretrained checkpoints.  We evaluate masked product token recovery as
an intrinsic probe, fine-tune on ESCI and WANDS for graded similarity and
query level ranking, and fine-tune on GLUE for retention.  Task specific
metrics are given in the subsections below; fine-tuning settings are summarized
next and detailed in Appendix~\ref{app:finetuning}.

\paragraph{Fine-tuning.}  Dense retrievers are fine-tuned on ESCI and
cross encoder rerankers on WANDS using the dataset-provided splits without
resampling, with HuggingFace \texttt{Trainer} (seed~42) and all reported scores
on the held out test split.  ESCI uses a Sentence-Transformers bi-encoder with
CoSENTLoss~\citep{huang2024cosent} on graded query product pairs; WANDS uses a
CrossEncoder trained with BCE on concatenated query product inputs.  For each
model we run a lightweight hyperparameter sweep over learning rate and epoch
count and report the best saved checkpoint (Appendix~\ref{app:finetuning}).

\subsection{Masked Product Token Recovery}

We follow the masked token recovery protocol of ModernBERT and FineFineWeb~\citep{map2024finefineweb}.
Product titles and descriptions are truncated to 128, 256 or 512 tokens; a
span aware mask covering 15\% of tokens is applied; and the model predicts the
original tokens at top-$k$ with $k\in\{1,3,5\}$.
Appendix~\ref{app:token_recovery} reports the full title and description
matrices. RexBERT-base consistently outperforms ModernBERT-base across all
block sizes and top-$k$ settings.

\subsection{Graded Query-Product Similarity}
\label{sec:semantic_similarity}

We evaluate whether domain adapted MLM encoders produce better query product
embeddings than general purpose retrieval models when relevance grade is used
as the similarity target.  Starting from each RexBERT checkpoint, we train an
embedding model on the English \emph{train} split of Amazon ESCI.  Each example
is a (query, product title) pair, truncated when needed to fit the encoder
context window.  We use the ESCI relevance grade as the supervision
signal for similarity by mapping Exact, Substitute, Complement and Irrelevant
to 1.0, 0.66, 0.33 and 0.0, respectively, and optimize with CoSENT
loss~\citep{huang2024cosent} so that cosine similarity in the embedding space
preserves these graded relations.  Evaluation reports Spearman correlation
between predicted cosine similarities and the grade derived target scores on
held out query product pairs.

\begin{figure*}[t]
  \centering
  \includegraphics[width=0.75\linewidth]{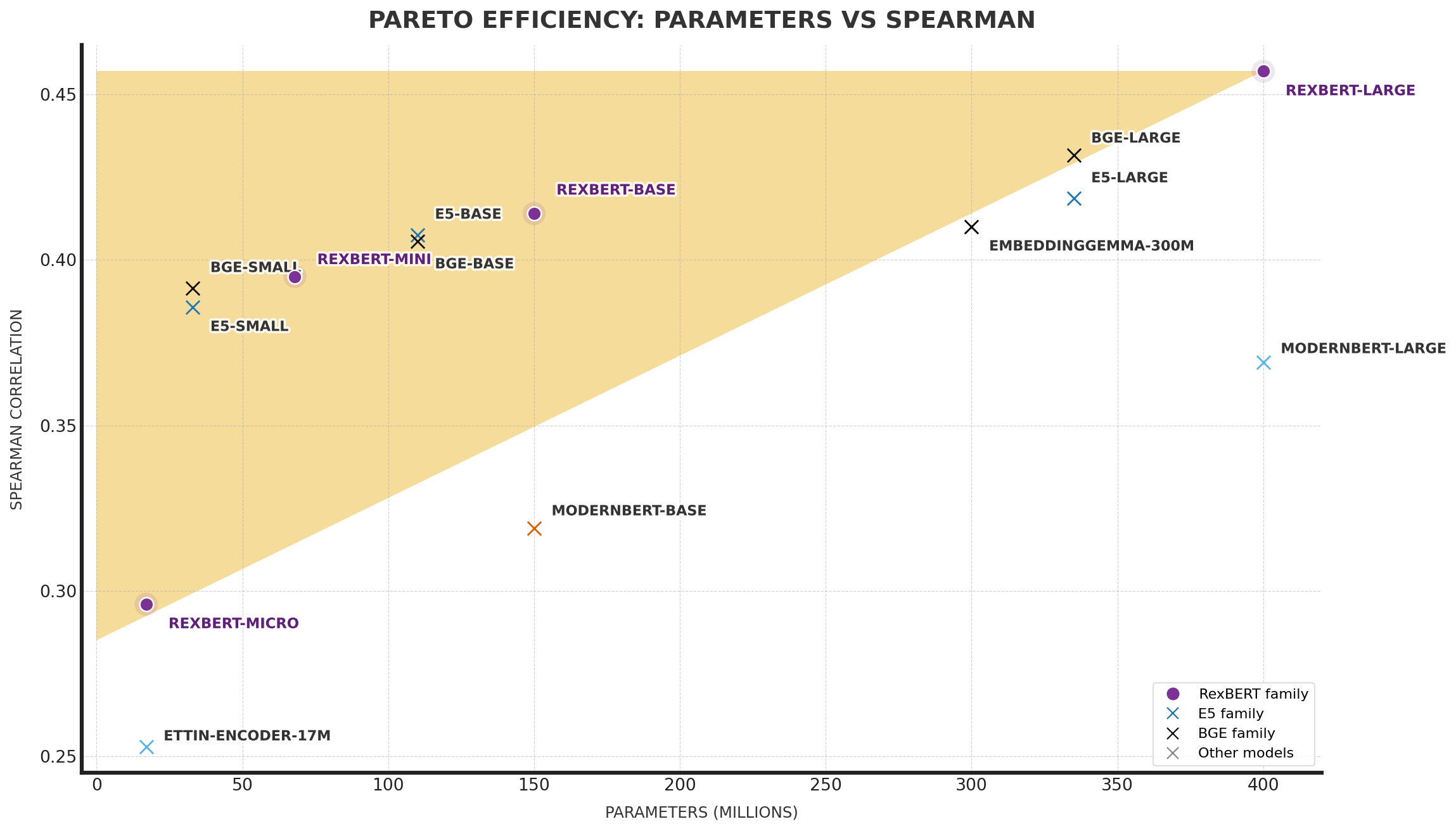}
  \caption{Pareto efficiency of ESCI graded similarity: Spearman correlation
  versus model size after embedding fine-tuning on the train split.  RexBERT
  models (Micro to Large) lie on the efficiency frontier and outperform
  Microsoft E5, BAAI BGE, ModernBERT, Ettin and EmbeddingGemma at comparable
  parameter scales.}
  \label{fig:similarity}
\end{figure*}

We compare RexBERT initialized embedding models against Microsoft E5 and BAAI
BGE encoders trained under the same ESCI protocol and metric.  E5 and BGE are
established retrieval embedding families that already undergo multiple stages of
contrastive pretraining before task specific fine-tuning; despite this, RexBERT
models still outperform them at comparable scale.  As shown in
Figure~\ref{fig:similarity} and Table~\ref{tab:embedding_baselines}, performance
improves from Micro through Large, with RexBERT-large reaching 0.457 Spearman,
suggesting that domain MLM pretraining transfers to stronger product relevance
embeddings when grade based similarity is the evaluation objective.

\begin{table}[t]
  \centering
  \small
  \caption{ESCI graded similarity after embedding fine-tuning on the train
  split.  Scores are Spearman correlations between predicted cosine similarity
  and grade derived targets on held out query product pairs.}
  \label{tab:embedding_baselines}
  \setlength{\tabcolsep}{3pt}
  \begin{tabular}{@{}llrc@{}}
    \toprule
    Family & Model & Params & Spearman \\
    \midrule
    RexBERT & Micro & 17M & 0.2991 \\
    RexBERT & Mini & 68M & 0.3955 \\
    RexBERT & Base & 150M & 0.4208 \\
    RexBERT & Large & 400M & 0.4571 \\
    \midrule
    Ettin & encoder-17m & 17M & 0.253 \\
    Ettin & encoder-68m & 68M & 0.302 \\
    \midrule
    ModernBERT & base & 150M & 0.319 \\
    ModernBERT & large & 400M & 0.369 \\
    \midrule
    EmbeddingGemma & 300m & 300M & 0.411 \\
    \midrule
    E5 & small & 33M & 0.3857 \\
    E5 & base & 110M & 0.4076 \\
    E5 & large & 335M & 0.4186 \\
    \midrule
    BGE & small & 33M & 0.3915 \\
    BGE & base & 110M & 0.4056 \\
    BGE & large & 335M & 0.4316 \\
    \bottomrule
\end{tabular}
\end{table}

\subsection{Product Search Relevance on WANDS}
\label{sec:wands_relevance}

WANDS~\citep{chen2022wands} provides 480 queries, 42{,}994 products and
233{,}448 human relevance judgments labeled \textit{Exact}, \textit{Partial} or
\textit{Irrelevant}.  Unlike the ESCI graded similarity task in
\S\ref{sec:semantic_similarity}, evaluation is query level ranking over each
query's labeled candidate products rather than pair level correlation.  We
fine-tune RexBERT, ModernBERT-base/large and BERT-mini/small on the WANDS train
split (dev used for checkpoint selection; Appendix~\ref{app:finetuning}),
representing each product by its name, description and structured catalog
features.

We also adapt Alibaba-NLP/gte-reranker-modernbert-base and
BAAI/bge-reranker-base as cross encoder baselines; both are already
strong rerankers trained on large scale relevance supervision before WANDS
fine-tuning.  At test time, models score every labeled query product pair,
candidates are ranked within each query, and we report NDCG@5, NDCG@10 and MAP
using \textit{Exact} $>$ \textit{Partial} $>$ \textit{Irrelevant} as graded gain.

Table~\ref{tab:wands_relevance} compares the RexBERT family against ModernBERT,
BERT and the GTE/BGE reranker baselines.  RexBERT models outperform the general
encoder and pretrained reranker baselines across NDCG@5, NDCG@10 and MAP, with
performance generally improving from Micro through Large.

\begin{table*}[t]
  \centering
  \small
  \caption{WANDS product search relevance after fine-tuning on the train query
  split and evaluation on held out test queries.  Bold marks the best score in
  each column.}
  \label{tab:wands_relevance}
  \setlength{\tabcolsep}{6pt}
  \begin{tabular*}{\textwidth}{@{\extracolsep{\fill}}llccc@{}}
    \toprule
    Model & Params & NDCG@5 & NDCG@10 & MAP \\
    \midrule
    \multicolumn{5}{@{}l@{}}{\textit{RexBERT}} \\
    Micro & 17M & 0.9122 & 0.9207 & 0.9722 \\
    Mini & 68M & 0.9157 & 0.9271 & 0.9739 \\
    Base & 150M & 0.9187 & 0.9314 & 0.9770 \\
    Large & 400M & 0.9239 & 0.9331 & 0.9774 \\
    \midrule
    \multicolumn{5}{@{}l@{}}{\textit{General encoder baselines}} \\
    ModernBERT-base & 150M & 0.9113 & 0.9243 & 0.9729 \\
    ModernBERT-large & 400M & 0.9198 & 0.9310 & 0.9749 \\
    BERT-mini & 11M & 0.8771 & 0.8962 & 0.9609 \\
    BERT-small & 29M & 0.8940 & 0.9108 & 0.9688 \\
    \midrule
    \multicolumn{5}{@{}l@{}}{\textit{Pretrained reranker baselines}} \\
    GTE reranker/ModernBERT-base & 150M & 0.9137 & 0.9271 & 0.9740 \\
    BGE reranker-base & 278M & 0.9127 & 0.9227 & 0.9725 \\
    \bottomrule
  \end{tabular*}
\end{table*}

\subsection{Guided MLM Ablation}
\label{sec:ablation_gmlm}

To isolate Guided MLM, we run a same-checkpoint ablation that varies only the
masking strategy while holding the Phase~2 initialization, data subset and token
budget fixed.  We evaluate the Base and Large ablation runs after 20B training
tokens using held out e-commerce perplexity.  Table~\ref{tab:gmlm_ppl_ablation}
shows consistent PPL reductions when Guided MLM is enabled.  Unlike the full
Ecom-niverse Phase~3 comparison in Table~\ref{tab:annealing_ppl}, this ablation
is restricted to 20B tokens so the effect of guided masking can be isolated.

\begin{table}[H]
  \centering
  \small
  \caption{Guided MLM perplexity ablation for Base and Large variants after
  20B training tokens, not the full Ecom-niverse run.  Lower PPL is better.}
  \label{tab:gmlm_ppl_ablation}
  \setlength{\tabcolsep}{3pt}
  \begin{tabular}{@{}lccc@{}}
    \toprule
    Variant & Guided MLM & PPL & $\Delta$ PPL \\
    \midrule
    Base & no & 5.67 & -- \\
    Base & yes & 5.14 & 0.53 \\
    Large & no & 5.32 & -- \\
    Large & yes & 4.68 & 0.64 \\
    \bottomrule
  \end{tabular}
\end{table}

\subsection{General NLU Retention}

We use GLUE as a retention check to verify that e-commerce specialization does
not catastrophically degrade general language understanding.  Table~\ref{tab:glue}
presents results across four model size categories.  These scores are not the
main claim of the paper; they show that the specialization gains on product
tasks do not come at the cost of broad collapse on standard NLU probes.

\begin{table*}[!t]
\caption{GLUE retention results comparing RexBERT models to contemporary
baselines across five tasks.  Bold indicates best performance within each
model size category.}
\label{tab:glue}
\centering
\small
\begin{tabular}{lcccccccc}
\toprule
& \multicolumn{2}{c}{\textit{Large}} & \multicolumn{2}{c}{\textit{Base}} & \multicolumn{2}{c}{\textit{Mini}} & \multicolumn{2}{c}{\textit{Micro}} \\
\cmidrule(lr){2-3}\cmidrule(lr){4-5}\cmidrule(lr){6-7}\cmidrule(lr){8-9}
Task & RexBERT & ModernBERT & RexBERT & ModernBERT & RexBERT & DistilBERT & RexBERT & ettin-17m \\
\midrule
SST-2 & 95.76 & \textbf{96.67} & 94.27 & \textbf{94.72} & \textbf{92.78} & 89.23 & \textbf{90.37} & 88.42 \\
QNLI & 94.00 & \textbf{94.82} & \textbf{92.75} & 90.88 & \textbf{91.51} & 87.99 & \textbf{85.59} & 85.30 \\
QQP & \textbf{89.84} & 89.73 & \textbf{89.28} & 89.26 & \textbf{88.52} & 87.42 & \textbf{85.90} & 85.54 \\
MNLI & \textbf{92.06} & 90.14 & 87.23 & \textbf{88.71} & \textbf{85.23} & 80.75 & \textbf{78.26} & 77.66 \\
STS-B & \textbf{91.80} & 91.33 & \textbf{89.55} & 85.92 & \textbf{88.53} & 84.84 & \textbf{83.00} & 82.62 \\
\bottomrule
\end{tabular}
\end{table*}

\section{Discussion}

The results in \S\ref{sec:experiments} show that when evaluation is strongly
e-commerce shaped, \emph{data and curriculum} dominate raw scaling: RexBERT
improves over general purpose encoders at comparable or larger parameter counts,
suggesting that representation quality in this domain is bottlenecked less by
model capacity and more by exposure to the right long tail entities,
attribute value constructions, and training distribution.

\paragraph{Why does specialization help?}
E-commerce language differs from generic web text in systematic ways: it is
entity dense, highly compositional, and often expressed in semi structured fragments. The \textbf{Ecom-niverse}
corpus is explicitly designed to cover this space, and the gains we observe
are most pronounced in settings that stress these properties. This supports the
hypothesis that the encoder benefits from learning (i) robust lexical
representations for domain terms and structure, and (ii)
contextual rules that bind attributes to the correct head entities.

\paragraph{Is annealing enough for adaptation?}
Because RexBERT starts from the Ettin Phase~2 checkpoint, its improvements are
attributable to the domain annealing stage rather than to full e-commerce
pretraining from scratch.  This is a practical advantage: many domains may not
have the compute budget to repeat general pretraining, but can afford a focused
distribution shift over high-quality in domain data.  Downstream transfer
(\S\ref{sec:experiments}) and held out e-commerce perplexity at the Base scale
provide the key evidence: comparing the unannealed Ettin Phase~2 checkpoint
with continued domain MLM and the final Phase~3 RexBERT checkpoint before any
task specific fine-tuning, perplexity drops monotonically from 6.04 to 4.93 to
3.99, showing that domain annealing shifts the inherited encoder toward
product language distributions.

\begin{table}[t]
  \centering
  \small
  \caption{Held out e-commerce perplexity comparison at the Base scale after
  full Ecom-niverse Phase~3 training.  Lower is better.}
  \label{tab:annealing_ppl}
  \begin{tabular}{@{}lcc@{}}
    \toprule
    Checkpoint & Guided MLM & PPL \\
    \midrule
    Ettin Phase~2 & no & 6.04 \\
    Domain MLM, no Guided MLM & no & 4.93 \\
    RexBERT Phase~3 & yes & 3.99 \\
    \bottomrule
  \end{tabular}
\end{table}

\paragraph{Guided MLM as a domain signal amplifier.}
Guided MLM concentrates MLM signal on rare but retrieval critical spans such as
product names, materials, sizes and colours.  The ablations in
\S\ref{sec:ablation_gmlm} and Table~\ref{tab:annealing_ppl} show that this bias
improves product language fit without replacing standard random masking
entirely.

\paragraph{Practical deployment implications.}
The observed gains translate directly to several production facing use
cases. First, stronger token recovery on titles/descriptions suggests
improved representations for attribute completion and normalization, which
can support catalog quality pipelines. Second, higher Spearman correlation
on graded relevance implies more faithful embedding geometry for retrieval
and candidate generation, especially in distinguishing \textit{substitute}
vs.\ \textit{complement} relations that generic models often conflate.

\section{Conclusion}

We presented \textbf{RexBERT}, domain adapted encoders for e-commerce product
understanding initialized from Ettin and specialized on \textbf{Ecom-niverse}
via Phase~3 annealing and \textbf{Guided MLM}.  Downstream evaluations show
consistent gains on product language and relevance benchmarks over general
encoders and task matched baselines, with competitive GLUE retention.  These
results support domain annealing as a practical recipe: inherit a strong
long context encoder, shift its training distribution toward curated in domain
data, and target salient product spans without repeating general pretraining
from scratch.  Future work includes broader multilingual coverage and tighter
contamination reporting.

\section*{Limitations}

RexBERT is evaluated primarily on English e-commerce data, so the results do
not establish comparable gains for other languages, marketplaces, regions or
time periods.  Product relevance can vary across sellers and geographies, and
category imbalance may cause the model to represent popular verticals better
than sparse ones.  The corpus curation pipeline depends on domain labels,
LLM assisted relevance judgments and classifier thresholds, each of which can
introduce coverage gaps or systematic filtering bias.  For reproducibility, we
release RexBERT checkpoints, Ecom-niverse artifacts, and Phase 3 pretraining code.

\makeatletter
\ifacl@finalcopy
\section*{Acknowledgments}

We gratefully acknowledge \citet{warner2025smarter} for
releasing ModernBERT and \citet{weller2025seq} for open sourcing the
Ettin encoder checkpoints.  RexBERT initializes from the public Ettin 8{,}192-token
Phase~2 checkpoint; the author-controlled training in this work is the
Ecom-niverse domain annealing phase.  We also thank \citet{map2024finefineweb} for
FineFineWeb and the HuggingFace community for
maintaining the infrastructure that enabled this work.
\fi
\makeatother

\bibliography{rexbert}
\clearpage
\appendix
\onecolumn
\section{Data and Corpus Details}
\label{app:data_details}

This appendix supplements \S\ref{sec:data}.  Table~\ref{tab:domains} lists
FineFineWeb~\citep{map2024finefineweb} domains kept directly when commerce-aligned, or filtered with
LLM relevance labels distilled into per domain fastText classifiers.  The
350B-token mixture is deduplicated, English-filtered, and decontaminated
against Amazon ESCI and WANDS before Phase~3 training.

\begin{table}[h]
  \centering
  \small
  \caption{Selected domains from FineFineWeb. The left block lists domains
  requiring filtering, while the right block lists domains with almost complete
  commerce overlap.}
  \label{tab:domains}
  \setlength{\tabcolsep}{4pt}
  \renewcommand{\arraystretch}{1.0}
  \begin{tabularx}{0.65\linewidth}{@{}>{\centering\arraybackslash}X@{\hspace{0.8em}}|@{\hspace{0.8em}}>{\centering\arraybackslash}X@{}}
    \toprule
    \textbf{Filter required} &
    \textbf{Directly selected} \\
    \midrule
    \begin{minipage}[t]{\linewidth}
      \vspace{0pt}
      \centering
      Hobby \\
      News \\
      Health \\
      Entertainment \\
      Travel \\
      Food \\
      Automotive \\
      Sports \\
      Music\,\&\,Dance \\
      \vspace{0.5\baselineskip}
    \end{minipage}
    &
    \begin{minipage}[t]{\linewidth}
      \vspace{0pt}
      \centering
      Fashion \\
      Beauty \\
      Celebrity \\
      Movie \\
      Photo \\
      Painting
    \end{minipage}
    \\
    \bottomrule
  \end{tabularx}
\end{table}

\section{Model Configurations}
\label{app:model_configs}

Table~\ref{tab:sizes} lists the RexBERT model configurations referenced in
\S\ref{sec:model}.

\begin{table}[H]
  \centering
  \caption{RexBERT model configurations.}
  \label{tab:sizes}
  \setlength{\tabcolsep}{6pt}
  \renewcommand{\arraystretch}{1.1}
  \begin{tabularx}{\linewidth}{@{}l *{4}{>{\centering\arraybackslash}X}@{}}
    \toprule
    Param. & Micro & Mini & Base & Large \\
    \midrule
    Parameters & 17M & 68M & 150M & 400M \\
    Layers & 7 & 19 & 22 & 28 \\
    Hidden & 256 & 512 & 768 & 1{,}024 \\
    FFN & 384 & 768 & 1{,}152 & 2{,}624 \\
    Heads & 4 & 8 & 12 & 16 \\
    Context & 8{,}192 & 8{,}192 & 8{,}192 & 8{,}192 \\
    LR & $3e{-}3$ & $3e{-}3$ & $8e{-}4$ & $5e{-}4$ \\
    WD & $3e{-}4$ & $3e{-}4$ & $1e{-}5$ & $1e{-}5$ \\
    \bottomrule
  \end{tabularx}
\end{table}

\section{Fine-tuning Details}
\label{app:finetuning}

This appendix supplements \S\ref{sec:experiments} with downstream fine-tuning
protocols for Amazon ESCI and WANDS.

\paragraph{Data splits.}
ESCI train and test pairs are taken from the public Hugging Face release
\url{https://huggingface.co/datasets/tasksource/esci}, retaining English
(\texttt{product\_locale}) examples with \texttt{small\_version=1}: 419{,}653
train and 181{,}701 test query product pairs ($\approx$70/30; no dev split is
provided).
WANDS train, dev, and test splits are used directly from the public Hugging
Face release \url{https://huggingface.co/datasets/napsternxg/wands} (60/20/20 by
query: 140{,}068 train, 46{,}690 dev, and 46{,}690 test judgments; 233{,}448
labeled pairs in total).
We do not resample or relabel either benchmark.

\paragraph{Amazon ESCI (bi-encoder).}
We train a Sentence-Transformers bi-encoder with CoSENTLoss on
(query, \texttt{product\_title}) pairs.  ESCI labels $\{$E, S, C, I$\}$ are
mapped to $\{$1.0, 0.66, 0.33, 0.0$\}$.  Optimization uses AdamW, a
\texttt{NO\_DUPLICATES} batch sampler, warmup ratio 0.1, bf16, and
\texttt{load\_best\_model\_at\_end=True} with \texttt{save\_strategy="epoch"}.
Evaluation on the test split reports Pearson and Spearman correlation between
predicted cosine similarity and the gold score.

\paragraph{WANDS (cross encoder).}
We train a CrossEncoder with \texttt{num\_labels=1} and
BinaryCrossEntropyLoss on concatenated (query, \texttt{product\_name}) pairs.
Relevance labels $\{$0, 1, 2$\}$ map to $\{$0.0, 0.5, 1.0$\}$.  Training uses
\texttt{max\_seq\_length=256}, warmup ratio 0.1, bf16, and
\texttt{flash\_attention\_2} for ModernBERT-family checkpoints.  At test time we
score all labeled pairs, rerank within each query, and report NDCG@10, NDCG@5,
MAP at relevance $\ge 1$, and Spearman on queries with $\ge 2$ distinct labels
($n{=}421$).

\paragraph{Hyperparameter search.}
Both setups use HuggingFace \texttt{Trainer} with \texttt{seed=42}.  We sweep
learning rate over $\{$1e-5, 2e-5, 3e-5, 5e-5$\}$ (plus 5e-6 for the ESCI large
model only) and epochs over $\{$2, 3, 4, 5, 8, 10$\}$.  Per-GPU batch size is
set by model capacity: 512/256/128 for bi-encoders and 128/64/32/16 for
cross encoders.  We report the best peak/saved checkpoint per model.

\section{Full Token Recovery Results}
\label{app:token_recovery}

This appendix reports the complete masked product token recovery results
referenced in \S\ref{sec:experiments}.  Table~\ref{tab:tc_title} gives
top-$k$ accuracy on product titles and Table~\ref{tab:tc_desc} on product
descriptions at context lengths of 128, 256 and 512 tokens.

\begin{table}[H]
\centering
\small
\caption{Masked product token recovery accuracy on product titles.}
\label{tab:tc_title}
\setlength{\tabcolsep}{4pt}
\begin{tabular}{@{}l*{9}{c}@{}}
\toprule
Block Size
& \multicolumn{3}{c}{128} & \multicolumn{3}{c}{256} & \multicolumn{3}{c}{512} \\
\cmidrule(lr){2-4}\cmidrule(lr){5-7}\cmidrule(lr){8-10}
Top-$k$ token Accuracy
& $k=1$ & $k=3$ & $k=5$ & $k=1$ & $k=3$ & $k=5$
& $k=1$ & $k=3$ & $k=5$ \\
\midrule
\multicolumn{10}{c}{\textit{Large Models}} \\
\midrule
RexBERT-large
& \textbf{72.0} & \textbf{82.3} & \textbf{85.1}
& \textbf{74.4} & \textbf{84.3} & \textbf{87.0}
& \textbf{76.1} & \textbf{85.6} & \textbf{88.2} \\
ModernBERT-large
& 65.0 & 75.0 & 78.2
& 68.4 & 78.1 & 81.1
& 69.8 & 79.4 & 82.3 \\
\midrule
\multicolumn{10}{c}{\textit{Base Models}} \\
\midrule
RexBERT-base
& \textbf{69.2} & \textbf{79.7} & \textbf{82.6}
& \textbf{71.4} & \textbf{81.6} & \textbf{84.5}
& \textbf{72.6} & \textbf{82.7} & \textbf{85.5} \\
ModernBERT-base
& 60.5 & 70.6 & 74.1
& 64.2 & 74.1 & 77.4
& 65.5 & 75.4 & 78.7 \\
\midrule
\multicolumn{10}{c}{\textit{Mini Models ($\sim$70M Parameters)}} \\
\midrule
RexBERT-mini
& \textbf{65.58} & \textbf{76.4}  & \textbf{79.67}
& \textbf{67.15} & \textbf{77.85} & \textbf{81}
& \textbf{66.75} & \textbf{77.62} & \textbf{80.81} \\
DistilBERT
& 34.88 & 43.99 & 48.53
& 33.48 & 42.57 & 42.27
& 29.41 & 37.78 & 42.49 \\
\midrule
\multicolumn{10}{c}{\textit{Micro Models (10 to 17M Parameters)}} \\
\midrule
RexBERT-micro
& \textbf{55.73} & \textbf{66.75} & \textbf{70.56}
& \textbf{55.49} & \textbf{66.57} & \textbf{70.44}
& \textbf{51.14} & \textbf{62.6}  & \textbf{66.78} \\
BERT-mini
& 39.94 & 47.94 & 51.47
& 43.85 & 51.94 & 55.38
& 46.09 & 54.18 & 57.57 \\
\bottomrule
\end{tabular}
\end{table}

\begin{table}[H]
\centering
\small
\caption{Masked product token recovery accuracy on product descriptions.}
\label{tab:tc_desc}
\setlength{\tabcolsep}{4pt}
\begin{tabular}{@{}l*{9}{c}@{}}
\toprule
Block Size
& \multicolumn{3}{c}{128} & \multicolumn{3}{c}{256} & \multicolumn{3}{c}{512} \\
\cmidrule(lr){2-4}\cmidrule(lr){5-7}\cmidrule(lr){8-10}
Top-$k$ token Accuracy
& $k=1$ & $k=3$ & $k=5$ & $k=1$ & $k=3$ & $k=5$
& $k=1$ & $k=3$ & $k=5$ \\
\midrule
\multicolumn{10}{c}{\textit{Large Models}} \\
\midrule
RexBERT-large
& \textbf{75.7} & \textbf{86.8} & \textbf{89.8}
& \textbf{77.9} & \textbf{88.4} & \textbf{91.1}
& \textbf{79.7} & \textbf{89.6} & \textbf{92.1} \\
ModernBERT-large
& 71.9 & 82.6 & 85.7
& 75.0 & 85.3 & 88.2
& 77.4 & 87.2 & 89.8 \\
\midrule
\multicolumn{10}{c}{\textit{Base Models}} \\
\midrule
RexBERT-base
& \textbf{73.1} & \textbf{84.6} & \textbf{87.8}
& \textbf{75.5} & \textbf{86.3} & \textbf{89.3}
& \textbf{77.4} & \textbf{87.7} & \textbf{90.4} \\
ModernBERT-base
& 67.8 & 78.8 & 82.2
& 71.2 & 81.8 & 85.0
& 74.0 & 84.1 & 87.0 \\
\midrule
\multicolumn{10}{c}{\textit{Mini Models ($\sim$70M Parameters)}} \\
\midrule
RexBERT-mini
& \textbf{68.26} & \textbf{80.39} & \textbf{83.93}
& \textbf{71.99} & \textbf{83.49} & \textbf{86.73}
& \textbf{74.28} & \textbf{85.21} & \textbf{88.22} \\
DistilBERT
& 48.19 & 59.93 & 64.58
& 49.4  & 61.39 & 66.08
& 49.1  & 61.32 & 66.16 \\
\midrule
\multicolumn{10}{c}{\textit{Micro Models (10 to 17M Parameters)}} \\
\midrule
RexBERT-micro
& \textbf{57.18} & \textbf{70.28} & \textbf{74.7}
& \textbf{61.63} & \textbf{74.16} & \textbf{78.24}
& \textbf{64.01} & \textbf{76.08} & \textbf{79.94} \\
BERT-mini
& 49.6  & 60.11 & 64.23
& 52.37 & 62.67 & 66.62
& 54.74 & 64.86 & 68.76 \\
\bottomrule
\end{tabular}
\end{table}

\end{document}